\documentclass{article}

\usepackage[final]{ARLET_2024}

\usepackage[utf8]{inputenc} 
\usepackage[T1]{fontenc}    
\usepackage{hyperref}       
\usepackage{url}            
\usepackage{booktabs}       
\usepackage{amsfonts}       
\usepackage{nicefrac}       
\usepackage{microtype}      
\usepackage{xcolor}         
\usepackage{amsmath}
\usepackage{graphicx}       
\usepackage{hyperref}
\usepackage{multirow}       
\usepackage{tablefootnote}
\usepackage{makecell}
\usepackage{algorithm, algpseudocode}
\graphicspath{ {./figures/} }
\usepackage{float}          
\usepackage{wrapfig}

\title{Efficient Offline Reinforcement Learning: \newline First Imitate, then Improve}

\author{%
  Adam Jelley\\
  University of Edinburgh\\
  \texttt{adam.jelley@ed.ac.uk} \\
  \And
    Trevor McInroe\\
  University of Edinburgh\\
  \texttt{trevor.mcinroe@ed.ac.uk} \\
  \AND
  Sam Devlin\\
  Microsoft Research Cambridge\\
  \texttt{sam.devlin@microsoft.com} 
  \And
  Amos Storkey\\
  University of Edinburgh\\
  \texttt{a.storkey@ed.ac.uk} \\
}

\begin{document}

\maketitle  
\vspace{0mm}
\begin{abstract}
Supervised imitation-based approaches are often favored over off-policy reinforcement learning approaches for learning policies offline, since their straightforward optimization objective makes them computationally efficient and stable to train. However, their performance is fundamentally limited by the behavior policy that collected the dataset. Off-policy reinforcement learning provides a promising approach for improving on the behavior policy, but training is often computationally inefficient and unstable due to temporal-difference bootstrapping. In this paper, we propose a best-of-both approach by pre-training with supervised learning before improving performance with off-policy reinforcement learning. Specifically, we demonstrate improved efficiency by pre-training an actor with behavior cloning and a critic with a supervised Monte-Carlo value error. We find that we are able to substantially improve the training time of popular off-policy algorithms on standard benchmarks, and also achieve greater stability. Code is available at: \url{https://github.com/AdamJelley/EfficientOfflineRL}.
\end{abstract}

\vspace{0mm}
\section{Introduction}



Supervised imitation-based approaches \citep{emmons_rvs_2022, chen_decision_2021, peng_advantage-weighted_2019} are becoming increasingly popular for learning policies offline due to their straightforward optimization objective and ease of use with large foundation models \citep{kim_openvla_2024, zhao_learning_2023, brohan_rt-2_2023, brohan_rt-1_2022}. However, recent work has highlighted the limitations of these supervised approaches and demonstrated that off-policy reinforcement learning (RL) leads to improved performance when the data is sub-optimal or noisy, and can be more reliable for stochastic environments and long-horizon tasks \citep{kumar_when_2022, brandfonbrener_when_2022, paster_you_2022}. Given these seemingly opposing approaches, it is natural to ask: can we get the best of both supervised learning and temporal-difference learning for offline reinforcement learning? Specifically, can we provide an approach that provides the training efficiency and stability of supervised learning, while still gaining the performance benefits of multi-step temporal difference learning? 

In this paper, we investigate such a supervised pre-training approach to obtain approximate behavior policies and value functions before attempting improvement with off-policy RL. While a naive approach might simply be to pre-train the policy with behavior cloning, this generally does not provide much benefit (as we will show), since it does not provide a value function, which off-policy RL algorithms require to improve the policy. Our core contribution is to demonstrate that pre-training a critic is essential to achieve efficient and stable improvement beyond the behavior policy performance. Our approach can be straightforwardly combined with existing off-policy reinforcement learning algorithms with minimal modification, to improve their training efficiency and stability.

The remainder of the paper is structured as follows. We begin by reviewing the relevant background for offline reinforcement learning. We then present a motivational example to provide intuition for why supervised pre-training can improve subsequent off-policy RL, before building on this example to provide a theoretical analysis of the expected efficiency improvement from pre-training. We generalize our procedure for entropy-regularized reinforcement learning, and provide algorithms for applying our approach in practice to modern offline RL algorithms. Finally, we demonstrate that the hypothesized efficiency gains hold in practice on standard offline RL benchmarks.


\section{Preliminaries and Related Work}\label{sec:relatedwork}


Online reinforcement learning (RL) involves an agent taking actions according to a policy $\pi$ to interact with a Markov Decision Process (MDP). An MDP can be defined by the tuple $(\mathcal{S}, \mathcal{A}, \mathcal{T}, r, d_0, \gamma)$ where $\mathcal{S}$ is the state space, $\mathcal{A}$ is the action space, $\mathcal{T}(s'|s,a)$ is the transition probability distribution, $r : \mathcal{S} \times \mathcal{A} \rightarrow \mathbb{R}$ is the reward function, $d_0$ is the distribution of initial states, and $\gamma \in (0,1]$ is a discount factor. The goal is generally to learn the policy $\pi^*$ that maximises the expected discounted returns: $\pi^*=\arg\max_{\pi} \mathbb{E}_{\pi, \mathcal{T}} \left[\sum_{t=0}^\infty \gamma^t r(s_t,a_t)\right]$. Offline RL poses the same goal, but the policy $\pi$ must be learned from a fixed dataset of interactions from a behavior policy $\pi_B$, without any additional data collection. This behavior policy $\pi_B$ is generally unknown and arbitrarily optimal, and may be a human policy, a hardcoded policy, another agent's policy, or some mixture of policies.

Perhaps the most straightforward approach for learning a policy from offline data is behavior cloning (BC) \citep{pomerleau_efficient_1991}. Since the training is supervised, convergence is relatively stable and efficient, but the learned policy can at best match the performance of the behavior policy, since behavior cloning does not utilize reward information, and online performance may be brittle due to accumulating errors taking the agent out-of-distribution (OOD) of known states \citep{ross_reduction_2011}.

What if we want to improve on the behavior policy? Behavior cloning variants such as BC-$k\%$ \citep{levine_offline_2020} and Advantage-Weighted Regression \citep{peng_advantage-weighted_2019, peters_reinforcement_2007} utilize reward information to preferentially clone the behavior data. More recently, conditioning the policy on desired returns or goals \citep{srivastava_training_2021, ma_how_2022} has seen some success with transformer based approaches \citep{chen_decision_2021, janner_offline_2021, carroll_unimask_2022}. While these behavior cloning variants are sometimes able to achieve generalization to greater returns at test-time than observed in the dataset \citep{brandfonbrener_when_2022, kumar_when_2022}, in general their performance is more limited than true reinforcement learning approaches which can more effectively use mechanisms such as trajectory stitching \citep{paster_you_2022, yang_dichotomy_2022}.

A more promising approach to improve on the behavior policy is to use off-policy reinforcement learning, usually in the form of actor-critic algorithms for continuous environments \citep{lillicrap_continuous_2019, silver_deterministic_2014}. 
However, naively taking the offline dataset as the replay buffer for an off-policy algorithm often leads to performance no better than random \citep{fujimoto_off-policy_2019}. This occurs because as the actor and critic are optimized for maximal return, they inevitably go out-of-distribution of the data \citep{chen_information-theoretic_2019}. Since there are no additional interactions to provide correcting feedback as in the online case, growing extrapolation errors cause erroneous values, amplified by temporal-difference bootstrapping, which can lead to \textit{policy collapse} \citep{luo_finetuning_2023}.

Most modifications of off-policy reinforcement learning algorithms for the offline setting involve regularization of either the actions or the values towards the provided dataset to prevent this out-of-distribution extrapolation \citep{fu_closer_2022}. TD3+BC \citep{fujimoto_minimalist_2021} modifies TD3 \citep{fujimoto_addressing_2018} by introducing a behavior cloning term to regularize the policy towards the behavior policy. Alternatively, CQL \citep{kumar_conservative_2020} modifies Q-learning to reduce the values for out-of-distribution actions to prevent positive extrapolation error. However, since regularization towards the behavior policy or values limits performance improvement \citep{moskovitz_tactical_2022}, recent approaches instead aim to capture out-of-distribution uncertainty \citep{wu_uncertainty_2021}. SAC-N and EDAC \citep{an_uncertainty-based_2021} use the minimum of an ensemble of critics to obtain value estimates that minimize positive extrapolation error (with EDAC introducing an additional diversification loss over SAC-N to reduce the required ensemble size), so optimization is less likely to cause policy collapse. 


While these off-policy RL approaches can lead to better performance than modified imitation approaches, their convergence can be inefficient due to the bootstrapping in the Bellman update used for learning values. It is well established that the Bellman error can be a poor proxy for the real value error, particularly when used for incomplete, off-policy datasets \citep{fujimoto_why_2022, patterson_generalized_2022, schulman_equivalence_2018}, which causes difficulties with using the Bellman error as the sole objective for training value functions offline.  In this work we attempt to address these difficulties by pre-training with a supervised objective, using Monte-Carlo (MC) return estimates. 

Monte-Carlo (MC) return estimates have a long history in online reinforcement learning for integrating information over longer horizons \citep{sutton_reinforcement_2018, he_learning_2016, ostrovski_count-based_2017, oh_self-imitation_2018, imani_off-policy_2019, wilcox_monte_2022}. In the offline context, \citet{geng_improving_2024} propose blending the MC return into the TD targets to achieve efficiency and stability benefits. While their motivation is similar, since they do not have a separate pre-training phase, the MC values associated with the behavior policy will become outdated as the policy improves, which can limit final policy performance. $Q$-Transformer \citep{chebotar_q-transformer_2023} similarly claimed improved training efficiency at scale by incorporating MC values into the Bellman target via a max operation (as proposed by \citet{wilcox_monte_2022}). The motivation is again comparable to ours, but since there is no separate pre-training phase, performance may be more affected by variance in the MC values, as we find in Appendix~\ref{app:qtrans}. Additionally, this modification is only a secondary contribution of $Q$-Transformer that is not extensively investigated in isolation. We aim to more comprehensively understand and justify the use of MC targets for improving the efficiency of offline RL in this work.


\section{Motivational Example}

\begin{figure}[h!]
\centering
  \vspace{0mm}
  \includegraphics[width=\textwidth]{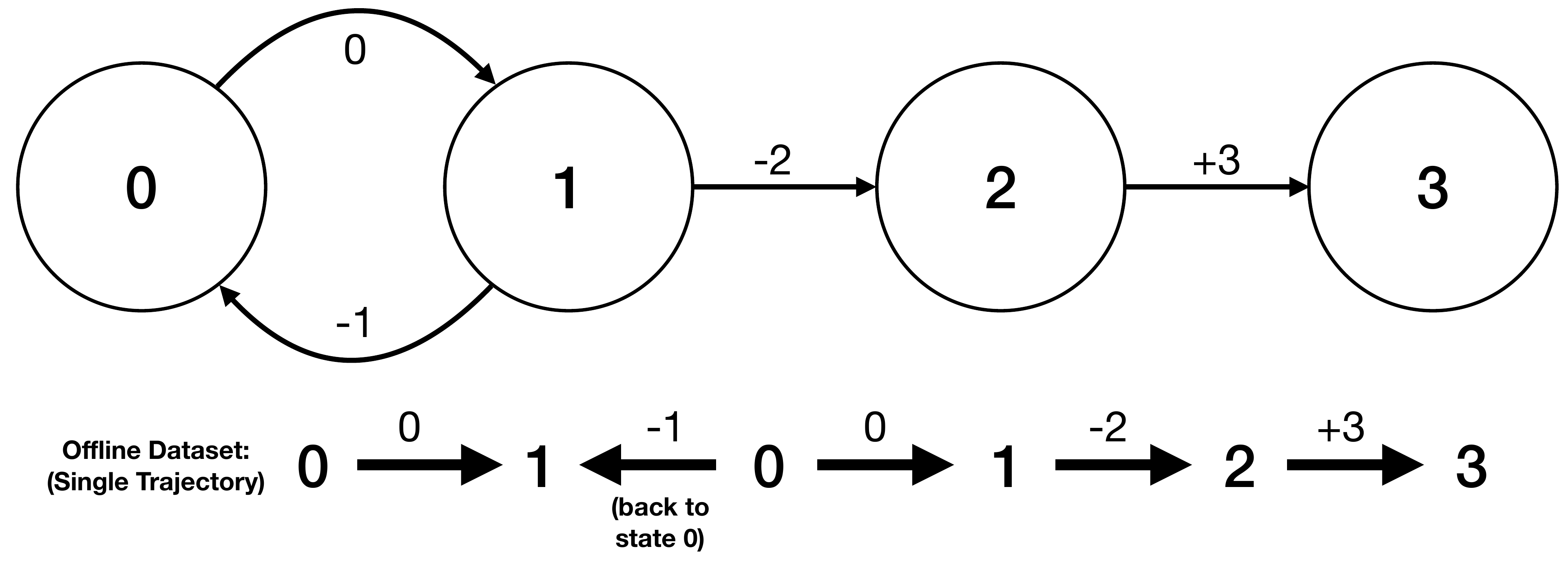}
  
  \caption[Figure 1 caption]%
  {A motivational tabular MDP. In offline reinforcement learning, we are provided with a dataset of trajectories. In this paper we utilize information from the entire trajectory (in the form of MC returns) to initialize a critic for subsequent off-policy reinforcement learning, which eliminates much of the inefficiency and instability associated with bootstrapping in temporal difference losses.}
\label{fig:MDP}
\vspace{0mm}
\end{figure}

As a motivational example, we consider offline tabular $Q$-learning on the simple 4 state MDP illustrated in Figure~\ref{fig:MDP}, together with a single offline trajectory from an unknown policy. In this example, we initialize all $Q$-values to zero (equivalent to randomly initializing a $Q$-network such that all initial $Q$-values are close to zero as is typical for deep $Q$-learning), and then perform temporal difference updates using the following update rule \citep{sutton_reinforcement_2018}:

\begin{equation}\label{eqn:Qlearning}
    Q(s_t,a_t) \leftarrow Q(s_t, a_t) + \alpha \left[r_t + \gamma \max_a Q(s_{t+1}, a) - Q(s_t, a_t)\right]
\end{equation}

For simplicity in this minimal motivational example, let us take the discount factor $\gamma$ and the learning rate $\alpha$ to be $1$. Performing updates using the provided offline trajectory leads to the $Q$-value updates shown in the left column of Table~\ref{tab:MDP}. We find that the policy converges to the optimal policy after 2 epochs, and the values converge to the correct optimal values after 3 epochs.

However, in the case of offline reinforcement learning, we have access to better initial $Q$-value estimates in the form of Monte-Carlo samples from the trajectory. This follows since in addition to Bellman's optimally equation in Equation~\ref{eqn:Qlearning}, we also have the definition of the Q-function for the behavior policy $\pi_B$ for which we can compute a sample-based expectation from our offline data:
\begin{equation}
    Q^{\pi_B}(s_t,a_t)=\mathbb{E}_{\tau \sim\pi_B}\left[\sum_{n=0}^{\infty}\gamma^{n} r_{t+n} \Big| s_t, a_t\right]
\end{equation}
Initializing the $Q$-values with Monte-Carlo estimates sampled from the provided trajectory and using the same update rule given in Equation~\ref{eqn:Qlearning} leads to the $Q$-values shown in the right column of Table~\ref{tab:MDP}. We now find that the policy converges to the optimal policy immediately after initialization, and the values converge to the optimal values after a single epoch. 

\begin{table}[h]
\caption{State-action values for each epoch of Q-learning to convergence for the motivational MDP and offline trajectory provided in Figure~\ref{fig:MDP} for both zero and Monte-Carlo value initializations.}
\label{tab:MDP}
\centering
\addtolength{\tabcolsep}{-1pt}
\begin{tabular}{| c | c c c c | c c c c |}
\hline
\hspace{-0.3em}\multirow{2}{2.2em}{Epoch}
& \multicolumn{4}{c|}{Zero Initialization} & \multicolumn{4}{c|}{MC Value Initialization}\\
\multicolumn{1}{|c|}{} & \multicolumn{1}{c}{$Q(0, \rightarrow)$} & \multicolumn{1}{c}{$Q(1, \leftarrow)$} &
\multicolumn{1}{c}{$Q(1, \rightarrow)$} &
\multicolumn{1}{c}{$Q(2, \rightarrow)$} & \multicolumn{1}{|c}{$Q(0, \rightarrow)$} & \multicolumn{1}{c}{$Q(1, \leftarrow)$} &
\multicolumn{1}{c}{$Q(1, \rightarrow)$} &
\multicolumn{1}{c|}{$Q(2, \rightarrow)$}\\
\hline
0 & 0 & 0 & 0 & 0 & 0 or +0.5\tablefootnote{The MC initialization value for $Q(0, \rightarrow)$ depends on the choice of first-visit or every-visit Monte-Carlo \citep{sutton_reinforcement_2018}. In this example it doesn't matter which is used after initialization.}  & \textbf{0} & \textbf{+1} & \textbf{+3}\\
1 & 0 & -1 & -2 & \textbf{+3} & \textbf{+1} & \textbf{0} & \textbf{+1} & \textbf{+3}\\
2 & -2 & -2 & \textbf{+1} & \textbf{+3} & \textbf{+1} & \textbf{0} & \textbf{+1} & \textbf{+3}\\
3 & \textbf{+1} & \textbf{0} & \textbf{+1} & \textbf{+3} & \textbf{+1} & \textbf{0} & \textbf{+1} & \textbf{+3}\\
\hline
\end{tabular}
\end{table}


This minimal example demonstrates the inefficiency of using uninformed initial values even in the absence of function approximation, near-optimal behavior policies, or complex MDPs. Equivalent inefficiency in bootstrapping from randomly initialized temporal difference (TD) targets will occur in the case of uninformed neural network initializations, particularly in sparse-reward environments where rewards received at the end of long trajectories may take many TD updates to propagate. 

\section{Theoretical Analysis}

In this section we provide a more formal analysis of the intuition provided by the minimal tabular $Q$-learning example above. We aim to investigate the functional form with which the $Q$-function initialization can affect the convergence to optimal $Q$-values. We consider the case of approximate fitted $Q$-iteration \citep{Riedmiller2005NeuralFQ}, in which:

\begin{equation}\label{eqn:Qiter}
\hat Q_{k+1} \leftarrow \arg\min_{\hat Q} ||\hat{Q}-\hat{T}\hat{Q}_k||_\infty
\end{equation}

where we have defined $\hat{Q}_k$ as the estimate of the $Q(s,a)$ function after $k$ iterations of Equation \ref{eqn:Qiter}, and $\hat T$ as the approximate Bellman operator $\hat{T}Q=\hat{r}+\gamma\hat{P}\max_a Q$, where $\hat r(s,a)$ and $\hat P(s'|s,a)$ are the sampled reward and environment transition probabilities respectively. The exact equivalents are denoted without the circumflex. We consider minimizing the infinite norm for each iteration since there is no convergence guarantee for the squared norm commonly used for empirical loss functions \citep{agarwal_reinforcement_2022}. In general, the error due to each $Q$-iteration (between successive iterations of the $\hat Q$ function) can be bounded as the sum of an `approximation' error, and a `sampling' error:

\begin{align}
    ||\hat Q_k - T \hat Q_{k-1}||_\infty &= ||\hat{Q}_k - \hat T \hat{Q}_{k-1} + \hat{T}\hat{Q}_{k-1}-T\hat{Q}_{k-1}||_\infty\\
    &\leq \underbrace{||\hat{Q}_k-\hat{T}\hat{Q}_{k-1}||_\infty}_{\text{approximation}} + \underbrace{||\hat{T}\hat{Q}_{k-1}-T\hat{Q}_{k-1}||_\infty}_{\text{sampling}}
\end{align}

Since the offline data is provided for offline reinforcement learning, the samples from which the Bellman operator can be approximated is fixed. We additionally assume here that the entire dataset is sampled uniformly to convergence in each iteration, such that the sampling error is constant and unaffected by $Q$-function initialization. We therefore focus on the approximation error, and assume this is finite at each iteration, $||\hat {Q}_{k+1} - T\hat{Q}_k||_\infty \leq ||\hat {Q}_{k+1} - \hat{T}\hat{Q}_k||_\infty \leq \epsilon_k$. Since we are interested in the effect of $Q$-function initialization on final optimality, we follow \citet{agarwal_reinforcement_2022} to write:

\begin{align}
    ||\hat{Q}_k-Q^*||_\infty &= ||\hat{Q}_k - T\hat{Q}_{k-1}+T\hat{Q}_{k-1}-Q^*||_\infty\\
    &= ||(\hat{Q}_k-T\hat{Q}_{k-1}) + (T\hat{Q}_{k-1}-T{Q^*})||_\infty\label{eqn:fixedpoint}\\
    &\leq ||\hat{Q}_k-T\hat{Q}_{k-1}||_\infty + ||T\hat{Q}_{k-1}-T{Q^*}||_\infty\\
    &\leq \epsilon_{k-1} + ||T\hat{Q}_{k-1}-T{Q^*}||_\infty\\
    &\leq \epsilon_{k-1} + \gamma ||\hat{Q}_{k-1} - Q^*||_\infty\label{eqn:contraction}\\
    &\leq \sum_{i=0}^{k-1}\gamma^i \epsilon_{k-i-1} + \gamma^k||\hat{Q}_0-Q^*||_\infty
\end{align}

using the fact that $Q^*$ is a fixed point of $T$ in Equation \ref{eqn:fixedpoint} and that $T$ is a $\gamma-$contraction in Equation~\ref{eqn:contraction}. 

This provides a measure of the (inifite-norm) Bellman error after $k$ iterations of fitted $Q$-iteration. We assume that for final convergence to the optimal policy, we would like to achieve an error over the offline dataset of order $\delta$. Therefore for two different intializations of $Q$, $Q_0$ and $Q'_0$, we assume that it takes $k$ and $k'$ iterations respectively to reach this error level, and make the bound tight:

\begin{minipage}{0.48\linewidth}
\begin{equation}\label{eqn:delta}
    \delta = \sum_{i=0}^{k-1}\gamma^i \epsilon_{k-i-1} + \gamma^k||\hat{Q}_0-Q^*||_\infty
\end{equation}
\end{minipage}
\hspace{5mm}
\begin{minipage}{0.48\linewidth}
\begin{equation}\label{eqn:deltaprime}
    \delta = \sum_{i=0}^{k'-1}\gamma^i \epsilon'_{k'-i-1} + \gamma^{k'}||\hat{Q'}_0-Q^*||_\infty
\end{equation}
\end{minipage}

We now make the strong assumption that the approximation error is in fact independent of the initialization and iteration step, such that $\epsilon_i=\epsilon'_j=\epsilon \ \forall \ i,j \in \mathbb{N}$, so that we can proceed to combine Equations \ref{eqn:delta} and \ref{eqn:deltaprime} as follows (assuming without loss of generality that $k > k'$):

\vspace{0mm}
\begin{align}
    \sum_{i=k'}^{k-1}\gamma^i \epsilon &= \gamma^{k'}||\hat{Q'}_0-Q^*||_\infty - \gamma^{k}||\hat{Q}_0-Q^*||_\infty\\
    \frac{\epsilon(\gamma^{k'} - \gamma^{k})}{1-\gamma} &= \gamma^{k'}||\hat{Q'}_0-Q^*||_\infty - \gamma^{k}||\hat{Q}_0-Q^*||_\infty\\
    k-k' &= \frac{1}{\ln\gamma}\ln\left(\frac{\frac{\epsilon}{1-\gamma} - ||\hat{Q'}_0-Q^*||_\infty}{\frac{\epsilon}{1-\gamma} - ||\hat{Q}_0-Q^*||_\infty}\right)\label{eqn:efficiency}
\end{align}

We see that $\hat Q'_0 \rightarrow \hat Q_0 \implies k'\rightarrow k$, i.e. a given initialization requires a given number of fitted $Q$-iterations to converge, as expected. Notice also the dependency on $1/\ln\gamma$, supporting our intuition that the efficiency improvement will generally be greatest for long-horizon tasks. We also see that as $\hat Q'_0$ becomes closer to $Q^*$ (in terms of the infinite norm over all states and actions), the numerator grows with respect to the denominator and so $k-k'$ increases logarithmically, demonstrating that the number of fitted $Q$-iterations $k'$ decreases as the $Q'_0$ initialization improves. 

In this work, we parameterize the $Q$ function with a network and assume this can be initialized to the approximate behavior value function $Q_B$ by pre-training the network with a supervised Monte-Carlo value error, so $\hat Q'_0=\hat Q_B$. If we assume that the behavior policy $\pi_B$ that generated the offline data achieves returns better than random (zero), we expect that $||\hat{Q}_B-Q^*||_\infty<||Q^*||_\infty$. If this holds, then Equation \ref{eqn:efficiency} demonstrates that the training efficiency (number of $Q$-iterations to convergence) should improve by initializing $\hat Q'_0=\hat Q_B$ compared to $\hat Q_0=0$. However, there are many ways in which this theory could be deviated from in practice, such as the use of generalized value iteration (rather than fitted $Q$-iteration) in modern deep actor-critic methods, the use of a squared- rather than infinite-norm loss function, the strong assumption that the error in each fitted $Q$-iteration is constant for any initialization, and the fact that variance in Monte-Carlo value targets may be such that $||\hat{Q}_B-Q^*||_\infty \not < ||Q^*||_\infty$ (especially given the infinite norm). Despite these limitations, this theoretical analysis provides clear motivation for our practical implementation outlined below.

\section{Pretraining Off-Policy Reinforcement Learning Algorithms in Practice}\label{sec:procedure}

\subsection{Outline Procedure}\label{sec:hardprocedure}

We now explain our pre-training procedure for standard return maximizing algorithms. In this case, we first compute the discounted return-to-go $R$ from each state-action pair until the end of the trajectory for all timesteps in the dataset. We can then pre-train the actor with behavior cloning and the critic ($Q$-network) with the pre-computed discounted return-to-go $R$, both using supervised mean-squared error (MSE) minimization (or alternatively cross-entropy for discretized actions/values). 
This procedure provides consistent initial actor and critic networks. Subsequently, a suitable off-policy RL algorithm (using a temporal-difference loss) can be applied to these pre-trained actor and critic networks to efficiently increase the policy return. Our pre-training procedure is outlined in pseudocode in Algorithm~\ref{alg:hardRL}. As we will see in Section~\ref{sec:MuJoCo}, this increase in training efficiency more than makes up for the time and computational expense associated with the supervised pre-training.

\subsection{Bias-Variance and Optimism-Pessimism Trade-offs}\label{subsec:biasvariance}
Under the behavior policy, this discounted return to go provides a Monte-Carlo (MC) sample of the expectation that the critic or $Q$-network aims to predict. In the case of deterministic environments and a single deterministic behavior policy, this Monte Carlo sample will equal the expectation exactly. For stochastic behavior policies and environments, this Monte Carlo sample may become high variance for finite sample datasets, which can lead to performance drops after pre-training. In this case it would be possible to use an $n$-step or $\lambda$-return to reduce this variance at the cost of bias introduced by bootstrapping (a well known case of the bias-variance trade-off) \citep{sutton_reinforcement_2018}. However, since computing the $\lambda$-return would require inferring the current critic value for every downstream state in the trajectory, an alternative way of controlling this trade-off is simply to compute the $TD(0)$ return and combine it with the MC return using a trade-off parameter $\lambda\in[0,1]$:
\begin{equation}\label{eqn:tildeR}
    \tilde R = (1-\lambda)R + \lambda (r+\gamma Q(s',a'))
\end{equation}
For large offline datasets with good environment coverage, where the greatest training efficiency gains are possible, there should be sufficient MC samples to reduce this variance to a manageable level for pre-training, so $\lambda$ can be small in order to utilize more information from the offline data (i.e. the full return-to-go). However, for smaller datasets capturing stochastic policies and environments, larger $\lambda$ may be beneficial. We investigate the empirical effect of varying $\lambda$ in our experiments.

Additionally, by pre-training on sampled returns with a symmetric error, the critic is equally likely to under- or over-estimate the values of out-of-distribution actions when optimizing the policy after pre-training, even in the deterministic case where the returns are exact. This overestimation can lead to policy collapse as discussed in Section~\ref{sec:relatedwork}. Therefore it can be helpful to add some value regularization $\mathcal{R}(Q(s,a))$ during pre-training, such as that introduced in CQL \citep{kumar_conservative_2020} to effectively lower-bound the $Q$ function. We find in our experiments that including some value regularization can be beneficial when the offline data is limited. This is reflected in Algorithm \ref{alg:hardRL}.

\subsection{Generalization to Maximum Entropy Off-Policy RL Algorithms}\label{subsec:softAlgs}

The maximum entropy RL framework, and particularly Soft Actor-Critic \citep{haarnoja_soft_2019} along with offline variants such as SAC-N and EDAC \citep{an_uncertainty-based_2021}, have recently become popular for their improved robustness and sample efficiency relative to the `hard' return maximization considered above. This `soft' RL framework involves maximizing the expected return alongside the entropy of the policy, balanced by a temperature parameter $\alpha$ \citep{ziebart_modeling_2010}:
\begin{equation}
    \pi^* = \arg\max_\pi\mathbb{E}_{\tau\sim\pi,\mathcal{T}}\left[\sum_{t=0}^\infty \gamma^t r_t + \alpha \mathcal{H}(\pi(\cdot|s_t))\right]
\end{equation}
where $\mathcal{H}(\pi(\cdot|s_t)) = \mathbb{E}_{\tilde a \sim \pi(\cdot|s_t)}\left[-\log(\pi(\tilde a|s))\right]$ is the entropy of the policy $\pi$ in state $s_t$.

This also modifies the definition of the state-action value functions as follows:
\begin{equation}\label{eqn:softrtg}
    Q(s_t,a_t)=\mathbb{E}_{\tau\sim\pi,\mathcal{T}}\left[\sum_{n=0}^\infty\gamma^{n}r_{t+n} + \alpha \sum_{n=1}^\infty\gamma^n \mathcal{H}(\pi(\cdot|s_{t+n}))\Big|s_t,a_t\right].
\end{equation}
Therefore, in order to pre-train value functions for the behavior policy as before we must modify the return-to-go to incorporate these entropy bonuses from every future timestep except the first. However, since these entropy bonuses depend on the current policy, we now separate our pre-training procedure above into two phases. First, we pretrain our policy with soft behavior cloning:
\begin{equation}\label{eqn:softBC}
    \mathcal{L}_{\pi_\theta} = \mathbb{E}_{s,a\sim D, \tilde a\sim\pi}\left[\alpha \log(\pi_\theta(\tilde a | s)) - \log(\pi_\theta(a|s))\right]
\end{equation}

This provides an approximate behavior policy with which we can compute Monte Carlo samples of Equation~\ref{eqn:softrtg} as soft returns-to-go, which can be used to augment the offline dataset as in Section~\ref{sec:hardprocedure}. These soft returns-to-go can then be used as the targets to pre-train the critic to achieve consistency with the soft behavior cloned policy, and providing a springboard initialization for a soft off-policy RL algorithm to efficiently improve the policy. The full pseudocode for this procedure is outlined in Algorithm~\ref{alg:softRL}, and a further discussion of the rational for this procedure is included in Appendix \ref{app:rational}.

\begin{minipage}{0.50\linewidth}
\begin{algorithm}[H]
\caption{Pre-training Hard Off-Policy RL}\label{alg:hardRL}
\begin{algorithmic}
\Require Dataset $D$ for use as replay buffer
\State Initialize $\pi_\theta$ and $Q_\phi$ parameters, $\theta$ and $\phi$
\For {each transition $(s_t, a_t, s_{t+1}, r_t) \in D$}
    \State Compute $R_t=\sum_{n=0}^{T-t} \gamma^n r_{t+n}$\newline \indent\Comment{Or $n$-step$/\lambda$-return/$\tilde R$ (eq. \ref{eqn:tildeR})}
    \State Append $R_t$ to transition: \newline\indent\indent\indent\indent $(s_t, a_t, s_{t+1}, r_t, R_t)$
\EndFor
\While {not converged} \Comment{\textbf{Pre-Training}}
    \State Sample batch $B=(s, a, s', r, R) \sim D$
    \State Update $\theta$ with behavior cloning:\newline \indent$\mathcal{L}_{\theta} = \mathbb{E}_B\left[(\pi(s) - a)^2\right]$
    \State Update $\phi$ with (generalized) return:\newline\indent$\mathcal{L}_{\phi} = \mathbb{E}_B\left[(Q(s,a) - R)^2\right](+\mathcal{R}(Q(s,a))$
    \newline\indent\Comment{Optional regularization $\mathcal{R}(Q(s,a))$}
    \State Update target networks (Polyak update):\newline\indent$\phi'\leftarrow\tau\phi'+(1-\tau)\phi$
\EndWhile
\While {$t<T$}\Comment{\textbf{Off-Policy RL}}
    \State Sample batch $(s, a, s', r, R) \sim D$
    \State Apply hard offline RL update to \newline\indent pre-trained $\pi$ and $Q$ to improve returns
\EndWhile
\end{algorithmic}
\end{algorithm}
\end{minipage}
\hspace{0.8pt}
\begin{minipage}{0.52\linewidth}
\begin{algorithm}[H]
\caption{Pre-training Soft Off-Policy RL}\label{alg:softRL}
\begin{algorithmic}
\Require Dataset $D$ for use as replay buffer
\State Initialize $\pi_\theta$ and $Q_\phi$ parameters, $\theta$ and $\phi$
\While {not converged} \Comment{\textbf{Actor Pre-Training}}
    \State Sample batch $B=(s, a, s', r) \sim D$
    \State Update $\theta$ with soft behavior cloning: \newline\indent$\mathcal{L}_{\theta} = \mathbb{E}_{B, \tilde a\sim\pi}\left[\alpha \log(\pi(\tilde a | s)) - \log(\pi(a|s))\right]$
\EndWhile
\For {transition $(s_t, a_t, s_{t+1}, r_t) \in D$}
    \State $R_t=\sum\limits_{n=0}^{T-t} \gamma^n r_{t+n} + \alpha \sum\limits_{n=1}^{T-t}\gamma^{n}\mathcal{H}(\pi(\cdot | s_{t+n}))$ 
    \newline\indent\Comment{Or $n$-step$/\lambda$-return/$\tilde R$ (eq. \ref{eqn:tildeR})}
    \State Append $R$ to transition: $(s, a, s', r, R)$
\EndFor
\While {not converged} \Comment{\textbf{Critic Pre-Training}}
    \State Sample batch $B=(s, a, s', r, R) \sim D$
    \State Update $\phi$ with (generalized) soft return:\newline\indent$\mathcal{L}_{\phi} = \mathbb{E}_B\left[(Q(s,a) - R)^2\right] (+\mathcal{R}(Q(s,a))$ 
    \State Update target networks: $\phi'\leftarrow\tau\phi'+(1-\tau)\phi$
\EndWhile
\While {$t<T$}\Comment{\textbf{Off-Policy RL}}
    \State Sample batch $(s, a, s', r, R) \sim D$
    \State Apply soft RL updates to pre-trained $\pi$ and $Q$
\EndWhile
\end{algorithmic}
\end{algorithm}
\vspace*{-4mm}
\end{minipage}

\section{Experiments on D4RL MuJoCo}\label{sec:MuJoCo}

\subsection{Implementation Details}\label{sec:mujocoprocedure}
To experimentally validate the hypothesized benefits of pre-training off-policy RL algorithms, we consider the standard D4RL MuJoCo benchmark \citep{fu_d4rl_2021}. We use the standard HalfCheetah, Hopper and Walker2d environments and the \textit{medium} dataset (a suboptimal policy with approximately $1/3$ of the performance of the expert) of 1M transitions, since this provides a meaningful behavior policy to learn from at initialization, with room for improvement with off-policy reinforcement learning. We also investigate the effect of pre-training on the \textit{medium-replay} and \textit{full-replay} datasets in Appendix~\ref{app:MRmujoco}, which also serve as sub-optimal but non-random behavior datasets.

For our implementations we utilize the Clean Offline Reinforcement Learning codebase (CORL) \citep{tarasov2022corl}, that provides algorithm implementations benchmarked to match published performance measures. During our experiments, we found that introducing LayerNorm \citep{ba_layer_2016} into both the actor and critic networks significantly improved training efficiency and stability, verifying the findings of \citet{ball_efficient_2023}. Therefore, for all results demonstrated in this paper, we use the benchmarked implementations from the CORL codebase, with the new addition of a LayerNorm after every linear layer except the final one for each network. We investigate the effect of LayerNorm in more detail in Appendix \ref{app:LN}.

We use TD3+BC \citep{fujimoto_minimalist_2021} as a `hard' off-policy algorithm, and EDAC \citep{an_uncertainty-based_2021} as a `soft' entropy-regularized algorithm to apply after pre-training. For EDAC, we include the auxiliary ensemble diversification loss as value regularization during pre-training, to prevent the collapse of the ensemble. We pre-train until convergence in each case, which can be determined by monitoring the convex supervised loss functions. Crucially, this corresponds to just 10-50k updates and is a small proportion of the total updates required for offline RL convergence. The online performance as a function of offline updates, including pre-training, is demonstrated below in Figure~\ref{fig:MuJoCo}. 

\subsection{Results and Analysis}
\label{sec:results}
\begin{figure}[h!]
\vspace*{0mm}
\centering
  \includegraphics[width=0.85\textwidth]{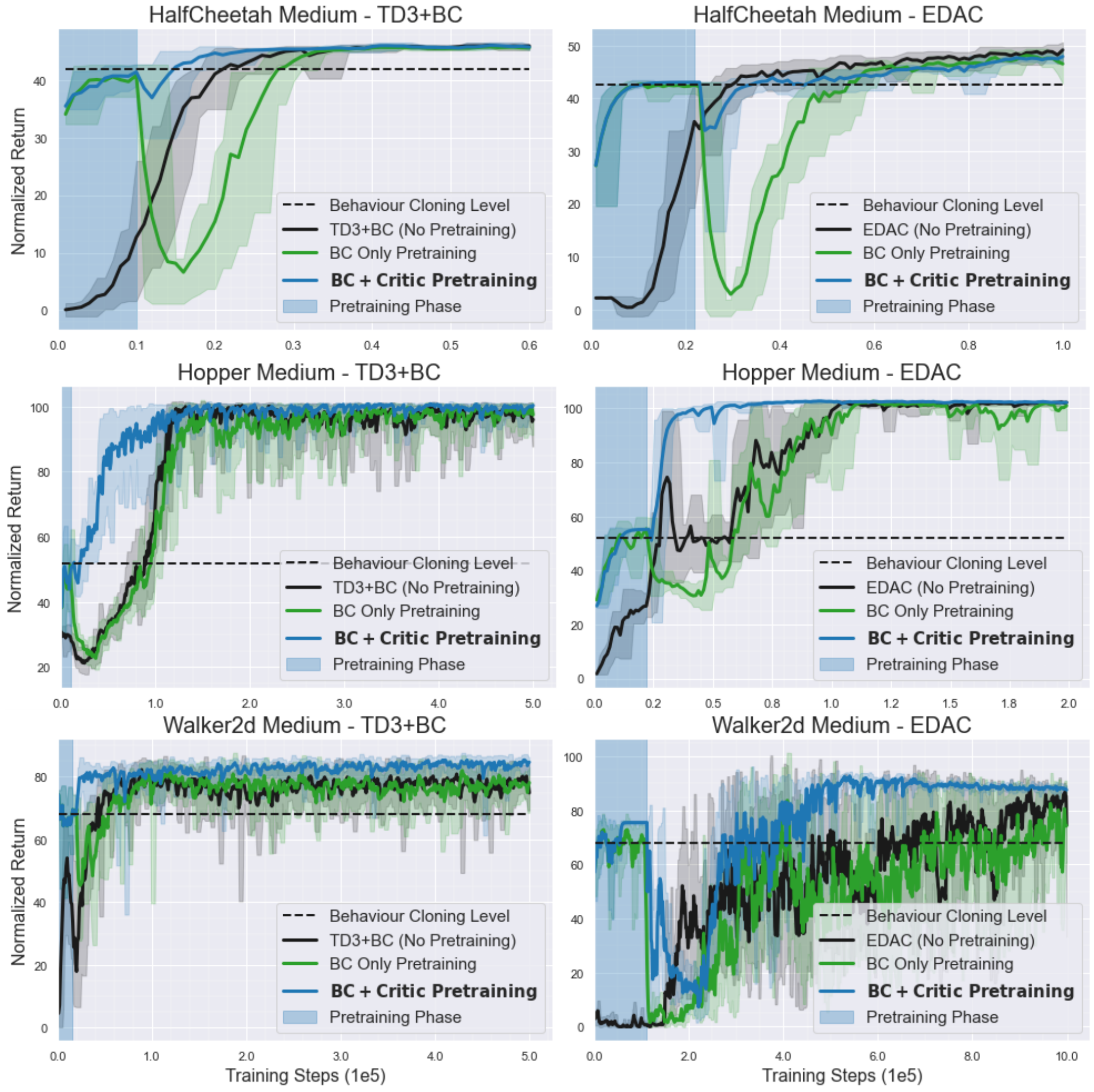}
  \vspace{0mm}
  
  \caption[Figure 2 caption]%
  {Performance means with shaded 95\% confidence intervals across 3 independent seeds. Supervised pre-training before offline reinforcement learning is more efficient than offline reinforcement learning from scratch. Surprisingly, performance is often more stable long after pre-training. }
 \vspace{0mm}
\label{fig:MuJoCo}
\end{figure}

We find that pre-training as described in Algorithms~\ref{alg:hardRL}, \ref{alg:softRL} leads to much more efficient training, both for TD3+BC (a hard RL algorithm with actor regularization) and for EDAC (a soft RL algorithm with critic regularization), even when taking the cost of pre-training into account. In particular, we find that on the more difficult Hopper and Walker2d environments, \textbf{the inclusion of pre-training  generally reaches asymptotic performance in less than $\mathbf{1/2}$ of the training steps and computation time required without pre-training}. Surprisingly, we notice that in many cases the final performance is also more stable, which we investigate further with ablations in Appendix \ref{app:Ablations}. We find similar benefits from pre-training on the \textit{medium-replay} and \textit{full-replay} datasets 
in Appendix~\ref{app:MRmujoco}.

In some cases, we found that the performance drops after critic pre-training. Fundamentally, this arises because at the end of pre-training we change the objective of the actor from imitation learning (predicting the action that would have been chosen in the dataset), to off-policy reinforcement learning (predicting the action that will maximize the critic prediction). If the values predicted by the critic are sufficiently accurate around the behavior policy after critic pre-training, the performance should smoothly improve, but if values are inaccurate due to variance then the performance may drop and take some time for Bellman updates to reduce this variance. In particular, the dynamics of these environments are deterministic, but the D4RL datasets are collected with a stochastic policy (SAC, \citet{haarnoja_soft_2018}) and from stochastic initial states, which leads to variance in the MC value estimates. Since the action spaces are larger for HalfCheetah and Walker2d than for Hopper (6 dimensional rather than 3), there will be more variance in the policy at each timestep, since the target policy entropy for SAC is the dimension of the actor space $\dim(A)$. This causes greater variance in the return. Furthermore, since all but one of the \textit{medium}-level Hopper trajectories end in termination, this increases the return signal-to-noise ratio for Hopper relative to the HalfCheetah and Walker2d environments which end with timeouts. To mitigate the performance drop off in these environments, we incorporate a small amount of TD target using Equation~\ref{eqn:tildeR}, with $\lambda=0.1$ for HalfCheetah, and Walker2d for EDAC. We investigate the effect of $\lambda$ fully in Appendix~\ref{app:lambda}. 

To further validate the benefit of pre-training, we additionally investigate the effect of pre-training on the Adroit environments \citep{rajeswaran_learning_2018} in Appendix~\ref{sec:adroit}. These are more complex and realistic tasks that require controlling a 24-DoF robotic hand to perform tasks such as aligning a pen, hammering a nail, opening a door, or relocating a ball. These form suitable environments for our approach, since the provided human demonstrations are much better than random behavior (and are therefore worth imitating) but can be improved upon by acting more efficiently, so it is possible to improve on this imitation policy. However, the human datasets provided for these environments are very limited, consisting of just 25 trajectories of human demonstrations per task. This means that additional regularization towards the behavior policy is necessary to prevent performance collapse after pre-training for the standard off-policy algorithms we consider. With this additional regularization in place, these experiments provide further evidence for the benefits of pre-training policies and values when learning a policy from effective but suboptimal human demonstrations.

\section{Conclusion}
We have demonstrated that pre-training off-policy algorithms to imitate the behavior policy and corresponding values to obtain a consistent actor and critic before attempting improvement can increase the computational efficiency and stability of subsequent off-policy reinforcement learning. In particular, we have shown that convergence should theoretically improve logarithmically as the pre-trained initialization improves (as measured by the infinite norm of the behavior policy values with respect to the optimal values). We have demonstrated experimentally that these efficiency improvements hold in practice for modern actor-critic algorithms, and can more than halve the computation time required for these algorithms to converge on standard offline reinforcement learning environments. Our proposed additional pre-training phase can be widely applied to current and future off-policy reinforcement learning algorithms and requires minimal algorithmic modification. We hope that our research contributes to bridging the gap between theoretically-tractable tabular algorithms and empirically driven deep reinforcement learning, while improving the efficiency and stability of offline reinforcement learning as scale continues to increase.


\subsubsection*{Broader Impact Statement}
\label{sec:broaderImpact}
Our work considers making offline reinforcement learning more computationally efficient and stable. The authors are not aware of any additional ethical concerns arising from our contributions that are not present in existing methods.

\appendix






\begin{ack}
Thank you to Eloi Alonso, Lukas Sch{\"a}fer and Tom Lee for insightful discussions. Adam Jelley was supported by Microsoft Research and EPSRC through Microsoft’s PhD Scholarship Programme.
\end{ack}

\newpage
\bibliography{references}
\bibliographystyle{apalike}


\appendix
\newpage

\section{Rational for Separation of Actor and Critic Pre-training for Entropy-Regularized Reinforcement Learning}\label{app:rational}

In Section \ref{subsec:softAlgs} and Algorithm \ref{alg:softRL} of the main text, we propose separating the pre-training of the policy and value network into two separate phases for entropy-regularized reinforcement learning algorithms. By first pre-training the policy with soft behavior cloning, an approximate behavior policy can be learned which then enables approximate behavior entropy bonuses to be included in the subsequent pre-training of the critic. However, an alternative approach could be to pre-train the policy and critic in parallel, as in the pure return maximization framework. In theory this would require updating the returns-to-go for each policy update to incorporate the changing entropy bonuses as the variance of the policy is updated. Since this requires a complete forward pass of the policy for all subsequent states in the trajectories of the states sampled for an update, this makes the pre-training significantly more computationally expensive. 

Another potential approach could be to only train the mean of the standard Gaussian policy to match the behavior policy, and keep the variance constant such that all entropy bonuses could be calculated at initialization and would be unaffected by policy pre-training. However, the standard tanh squashing applied to the policy to keep the sampled action within the environment action bounds leads to a changing entropy of the resulting policy, even with the Gaussian variance kept constant. 

A third approach could be to separate the policy and value pre-training, but continue the policy behavior cloning during value pre-training. While this is possible, for the entropy bonuses to be sufficiently accurate, the policy behavior cloning must have effectively converged before value pre-training, meaning continuing to pre-train the policy in parallel with the critic has little effect and requires unnecessary computation time. 

A final approach could be to compute the soft returns-to-go based on the initialization policy, and then only pre-train the values (no behavior cloning). While this approach was successful and led to training efficiency gains, the rapid updating of the actor at the beginning of training (and particularly the rapidly changing policy entropy) quickly leads to inconsistent values, so we found that the investment in pre-training the policy with soft behaviour cloning first was worth the computational time in most cases.
 \newpage
\section{Investigation Into Affect of LayerNorm}\label{app:LN}
\begin{figure}[h!]
\centering
  \includegraphics[width=\textwidth]{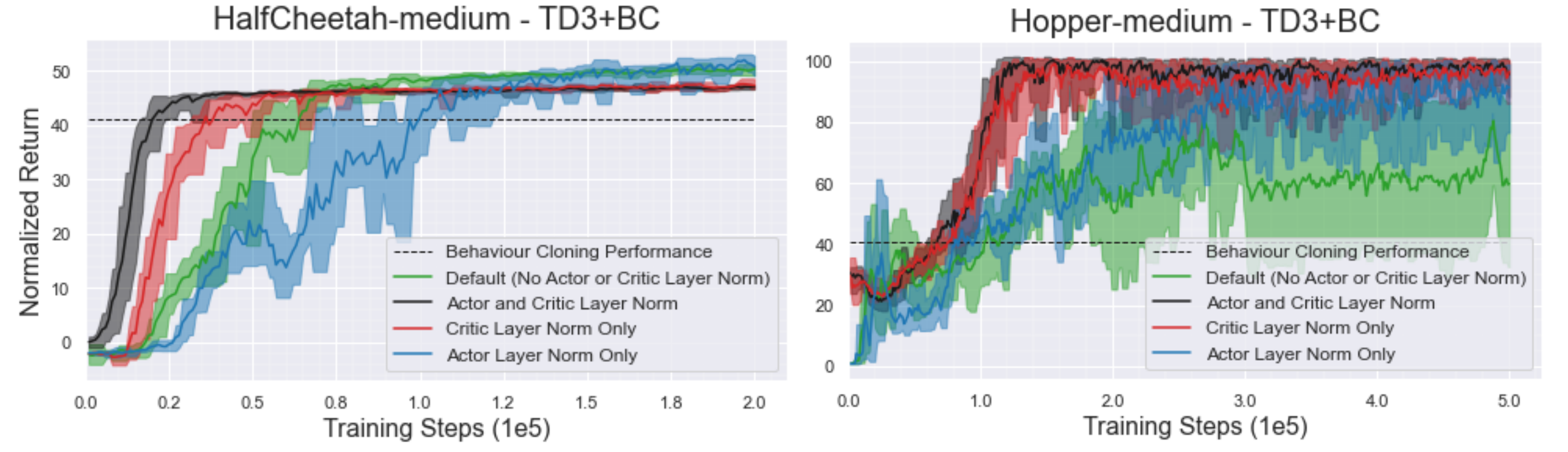}
  \vspace{-6mm}
  
  \caption[LayerNorm figure caption]%
  {Investigation into the affect of adding LayerNorm to both the actor and critic networks for TD3+BC on HalfCheetah and Hopper-medium. All lines show mean and standard deviation in normalized return at each timestep over 3 seeds.}
\label{fig:LayerNorm}
\end{figure}

We investigate the effect of the addition of LayerNorm \citep{ba_layer_2016} to the actor and critic networks individually for TD3+BC on the HalfCheetah-medium and Hopper-medium datasets. The original author implementation of TD3+BC (along with that of SAC-N, EDAC and many other RL algorithms \citep{tarasov2022corl}) does not include any form of normalization, and is shown in green. We consider the addition of LayerNorm after every linear layer in the network (before activation) except the final linear layer. We find that adding LayerNorms to the critic network leads to significant improvement in training efficiency and stability. This independently verifies the findings of \citet{ball_efficient_2023}, who hypothesize that this occurs because the normalization prevents severe value extrapolation for out-of-distribution actions, leading to overestimation error.
Surprisingly, the addition of LayerNorm to the actor alone leads to worse efficiency and stability than no LayerNorm for the HalfCheetah environment. However, the addition of LayerNorm to both the actor \textit{and} the critic leads to greater training efficiency and stability than the default, for both environments. 

We find that these insights generally hold across investigated environments and datasets, as shown in Figure~\ref{fig:FullLayerNorm} below. We therefore apply LayerNorm to both the actor and the critic for all experiments in this paper towards our goal of improving training efficiency. We expect the addition of LayerNorm to be universally used for off-policy reinforcement learning algorithms going forwards.

\vspace{0mm}
\begin{figure}[h!]
\includegraphics[width=\textwidth]{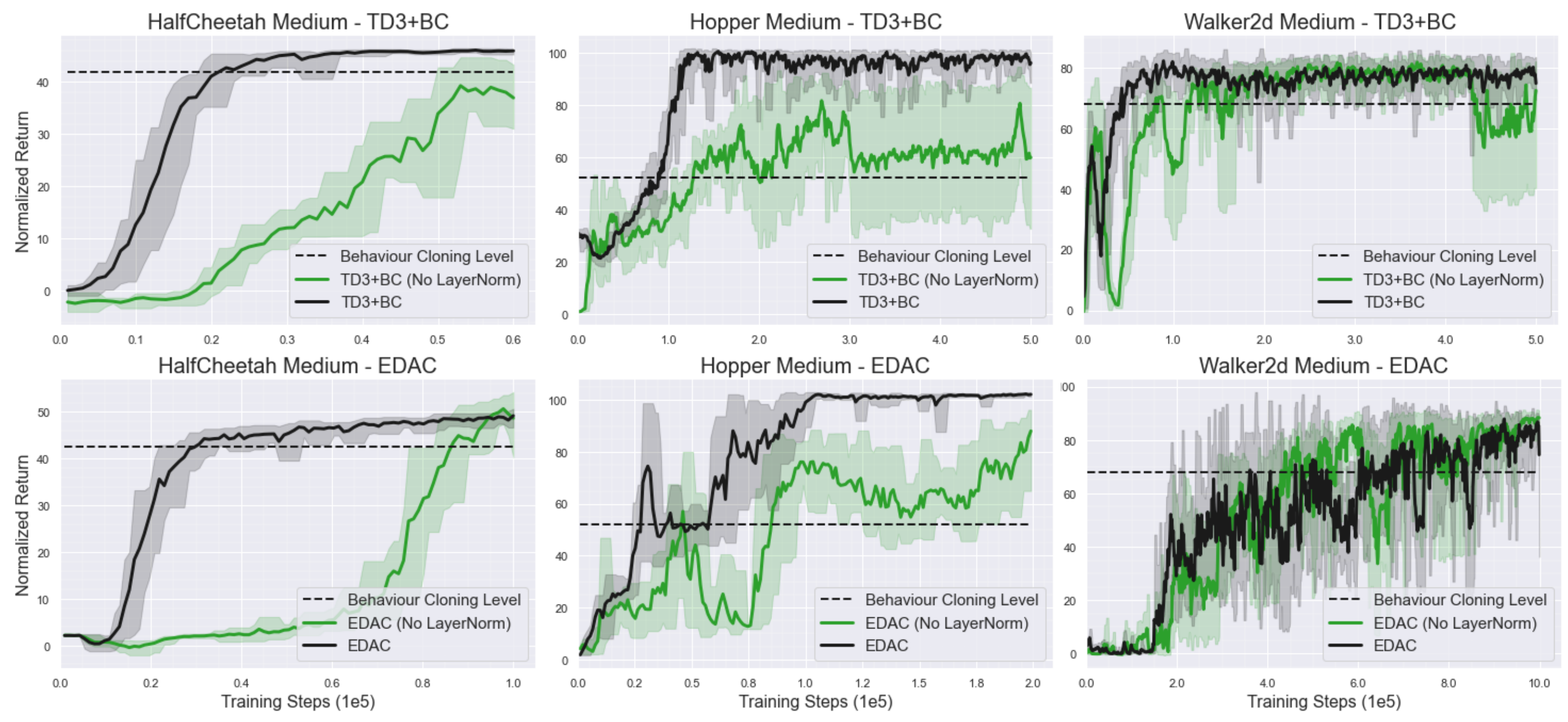}
  
  \caption[LayerNorm figure caption]%
  {The effect of adding LayerNorm to both actor+critic across environments. All lines show mean and standard deviation in normalized return at each timestep over 3 seeds.}
\label{fig:FullLayerNorm}
\end{figure}

\newpage
\section{Ablations of Actor and Critic Pre-Training}\label{app:Ablations}
\begin{figure}[h!]
\centering
\vspace{-5mm}  \includegraphics[width=\textwidth]{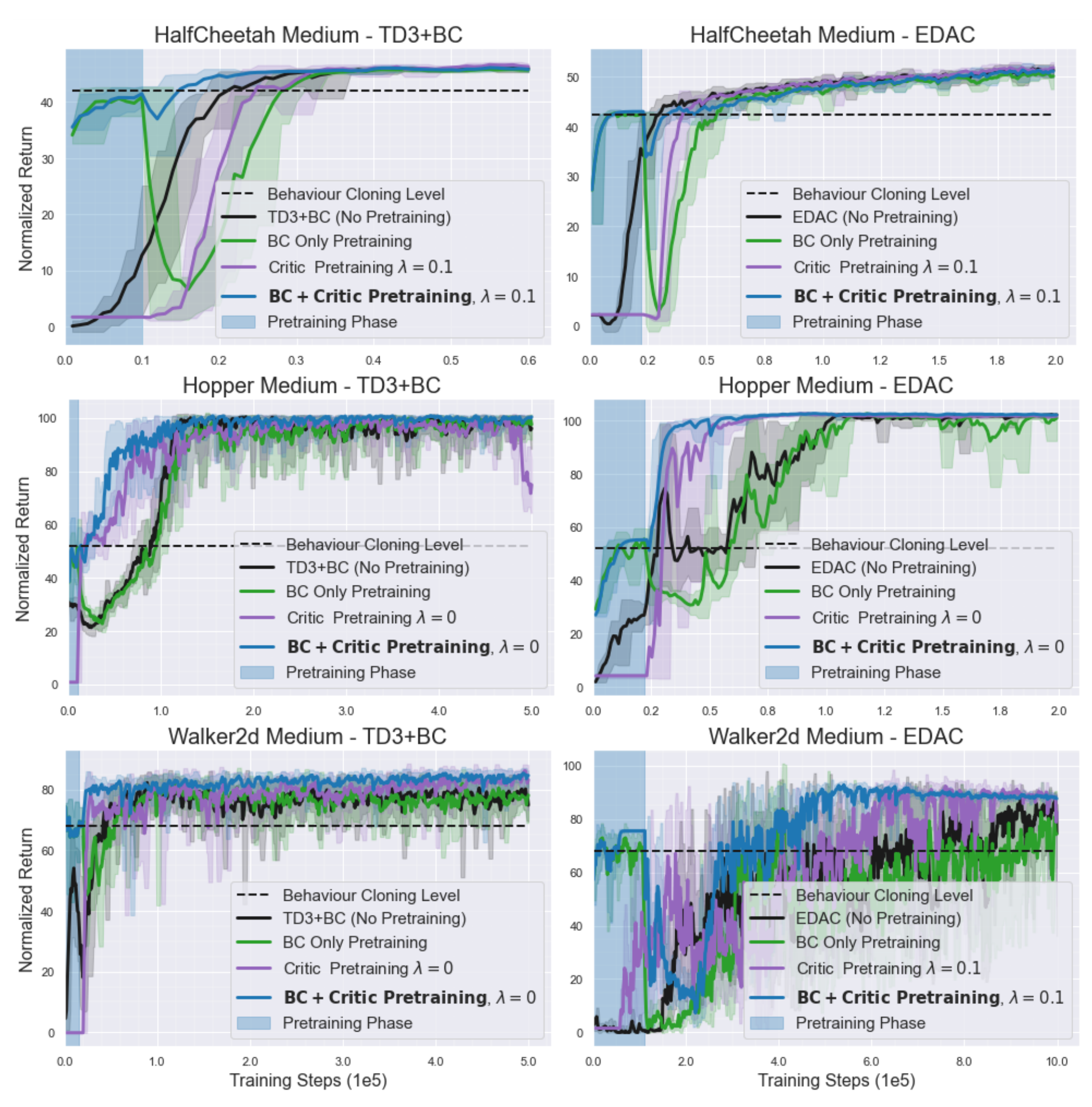}
  
  \caption[Ablation figure caption]%
  {Ablation of actor and critic pre-training. All lines show mean and standard deviation in normalized return at each timestep over 3 seeds.}
\label{fig:LayerNorm}
\end{figure}

We see that when only pre-training the actor with behavior cloning, the initial performance matches our proposed actor and critic pre-training, but quickly declines after pre-training due to the randomly initialized critic, and is no better than training from scratch. Pre-training the critic is much more important for efficient performance improvement, and is better than no pre-training, but performance starts from random so is not as effective as combined actor and critic pre-training. This demonstrates that having a consistent actor and critic is important for improved training efficiency and stability. 

Surprisingly, we also notice that in many cases the asymptotic performance with combined actor and critic pre-training is more stable, even after hundreds of thousands of updates after pre-training. As analyzed in \citet{fujimoto_why_2022} and \citet{chen_information-theoretic_2019}, for finite data regimes such as the offline setting, the Bellman equation can be satisfied by infinitely many suboptimal solutions. 
We hypothesize that this additional stability could be occurring because the initial pre-training to obtain a consistent actor and critic reduces the subset of possible solutions to those with lower value error when subsequently minimizing the Bellman error on the finite offline dataset. However, since this benefit is auxiliary to our central focus of improving efficiency, we leave investigation of this effect to further work.

\newpage
\section{Investigation into Empirical Bias Variance Tradeoff}\label{app:lambda}
\begin{figure}[h!]
\centering
  \vspace{-5mm}
  \includegraphics[width=0.8\textwidth]{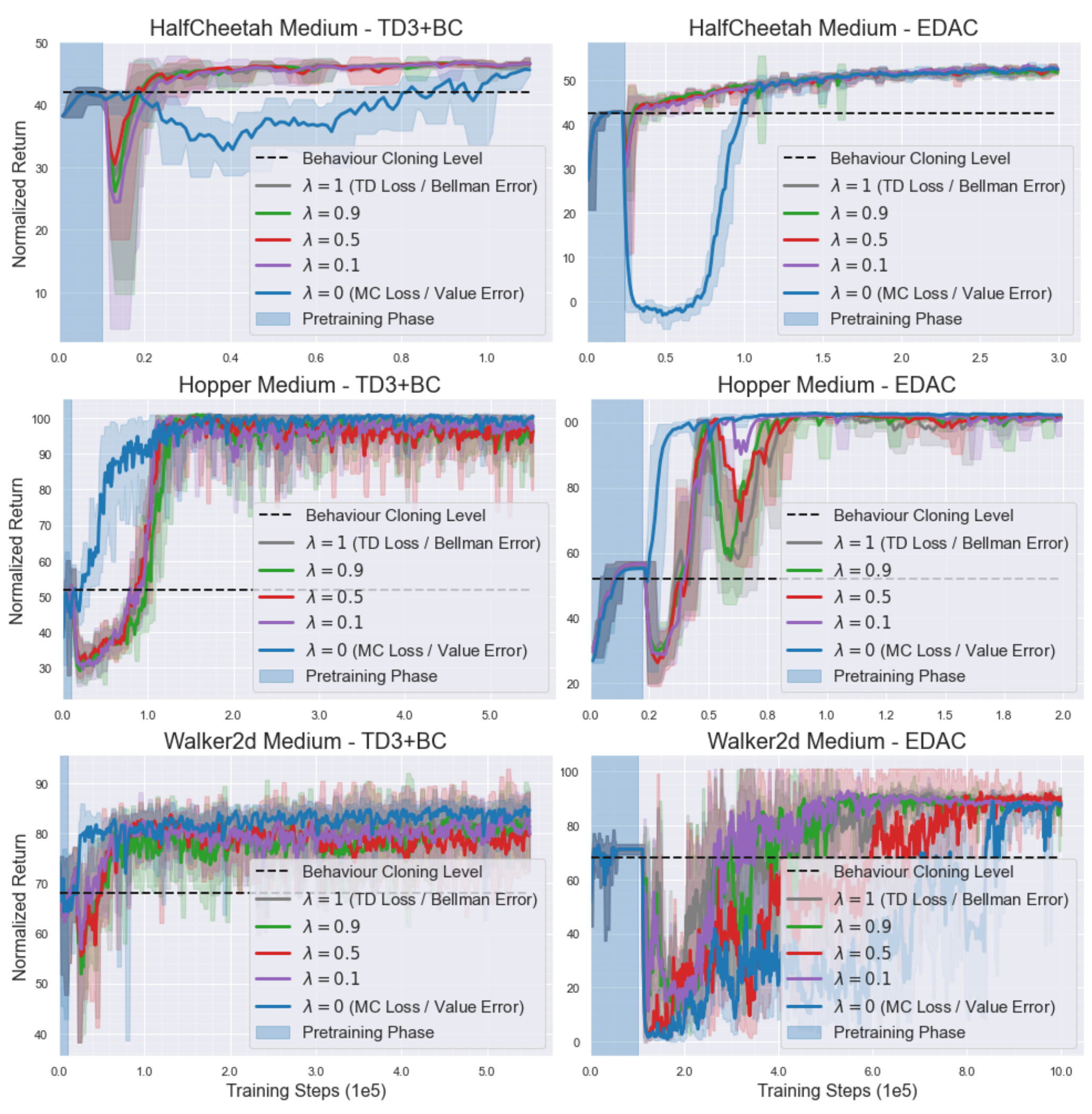}
  
  \caption[Lambda figure caption]%
  {Investigation into the empirical bias-variance tradeoff by varying $\lambda$ defined in Equation~\ref{eqn:tildeR}. All lines show mean and standard deviation in normalized return at each timestep over 3 seeds.}
\label{fig:Lambda}
\end{figure}
We investigate the bias-variance trade-off described in Section \ref{subsec:biasvariance} in practice by empirically varying $\lambda$ across algorithms and environments. We find that while all values of $\lambda \in [0,1]$ provide efficiency benefits over no pre-training, the effect of $\lambda=0$ (corresponding to fully supervised value-error pre-training) is often quite different, likely because even for $\lambda=0.1$ the temporal difference component to the loss can have significant impact on the training dynamics, and the pre-training duration is not long enough for this bootstrapping loss to reach consistency. However, including some temporal difference component ($\lambda>0$) can help to smooth the transition between imitation learning and off-policy reinforcement learning, as we see particularly evidently for the HalfCheetah environment. 

As discussed in the main text in Section \ref{sec:MuJoCo}, if the values predicted by the critic are sufficiently accurate around the behavior policy after critic pre-training, then the performance should smoothly improve, but if values are inaccurate due to variance or extrapolation error, then the performance may drop and take some time for Bellman updates to reduce this variance. In particular, the dynamics of these environments are deterministic, but the D4RL datasets are collected with a stochastic policy (SAC, \citet{haarnoja_soft_2018}) and from stochastic initial states, which leads to variance in the MC value estimates. Since the action spaces are larger for HalfCheetah and Walker2d than for Hopper (6 dimensional rather than 3), there will be more variance in the policy at each timestep, since the target policy entropy for SAC is the dimension of the actor space $\dim(A)$. This causes greater variance in the return. Furthermore, since all but one of the \textit{medium}-level Hopper trajectories end in termination, this increases the return signal-to-variance ratio for Hopper relative to the HalfCheetah and Walker2d environments, which end with timeouts. More generally, we expect that larger scale datasets with broad coverage of the relevant state-action space to have less variance in the MC return targets and therefore have less need for a TD component ($\lambda > 0$).
\newpage

\section{MuJoCo Medium-Replay and Full-Replay Dataset Experiments}\label{app:MRmujoco}
We apply our pre-training approach proposed in Section \ref{sec:procedure} with an identical experimental implementation to that described in Section \ref{sec:mujocoprocedure} to the \textit{medium-replay} and \textit{full-replay} datasets, shown below in Figures \ref{fig:MRMuJoCo} and \ref{fig:FRMuJoCo} respectively. 

\subsection{Medium-replay}
The \textit{medium-replay} dataset consists of 1M transitions from the replay buffer of an agent trained from random to medium performance. The effect of combined actor and critic pre-training on these datasets is shown below.
\begin{figure}[h!]
\vspace*{0mm}
\centering
  \includegraphics[width=\textwidth]{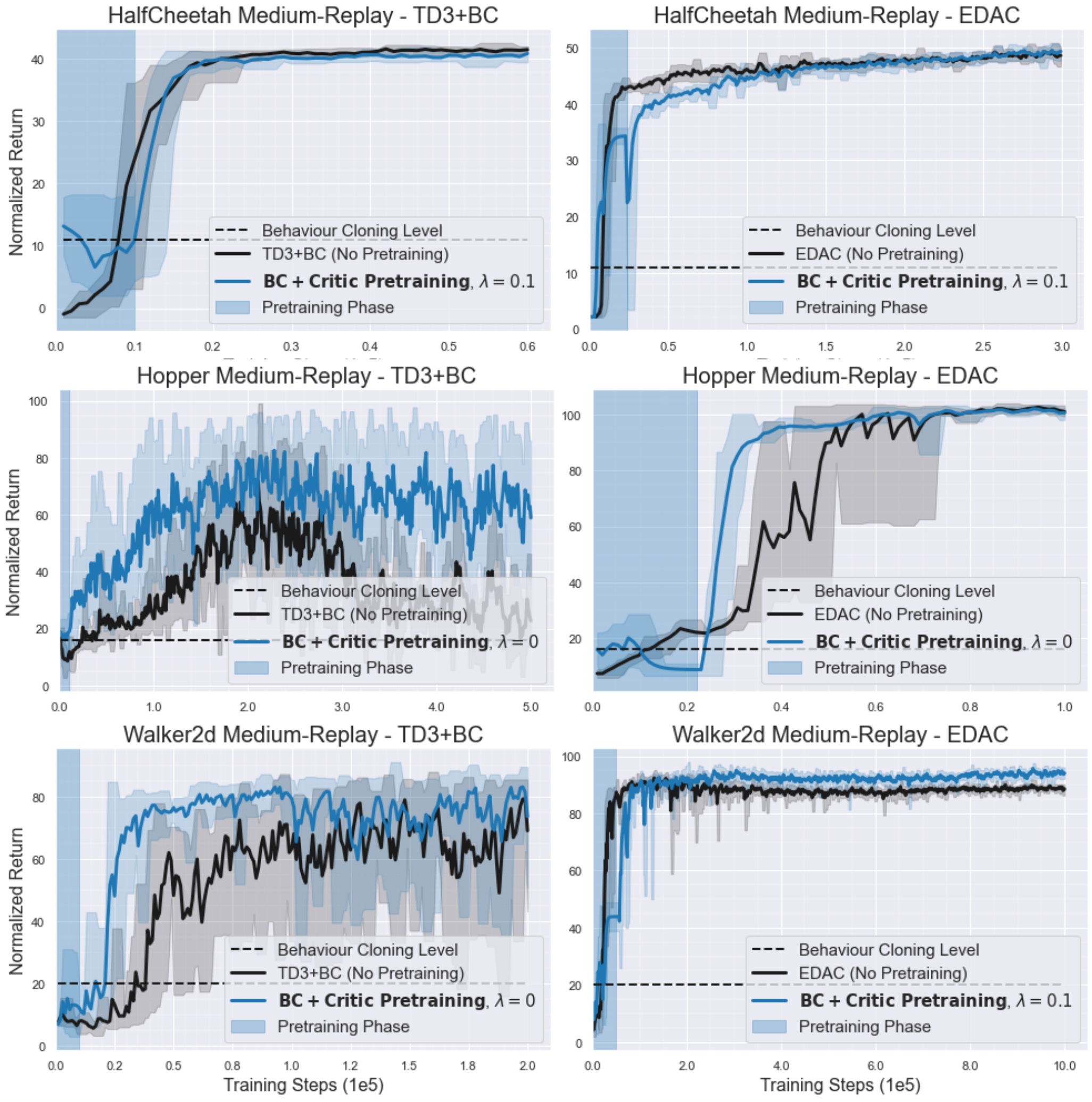}
  
  \caption[MR MuJoCo caption]%
  {Application of supervised pre-training to the \textit{medium-replay} MuJoCo datasets. We generally see training efficiency gains similar to those observed in Figure \ref{fig:MuJoCo}. Plots show mean and standard deviation at each timestep for 3 independent runs.}
\vspace*{0mm}
\label{fig:MRMuJoCo}
\end{figure}

We generally see similar efficiency and stability gains to those observed for the \textit{medium} datasets in Figure \ref{fig:MuJoCo}. We also notice that the performance of Walker2d for EDAC is much cleaner, likely due to greater diversity of data helping to stabilize performance.

\subsection{Full-replay}
The \textit{full-replay} dataset consists of 1M transitions from the replay buffer of an agent trained from random to expert performance. The effect of combined actor and critic pre-training on these datasets is shown below.
\begin{figure}[h!]
\vspace*{0mm}
\centering
  \includegraphics[width=\textwidth]{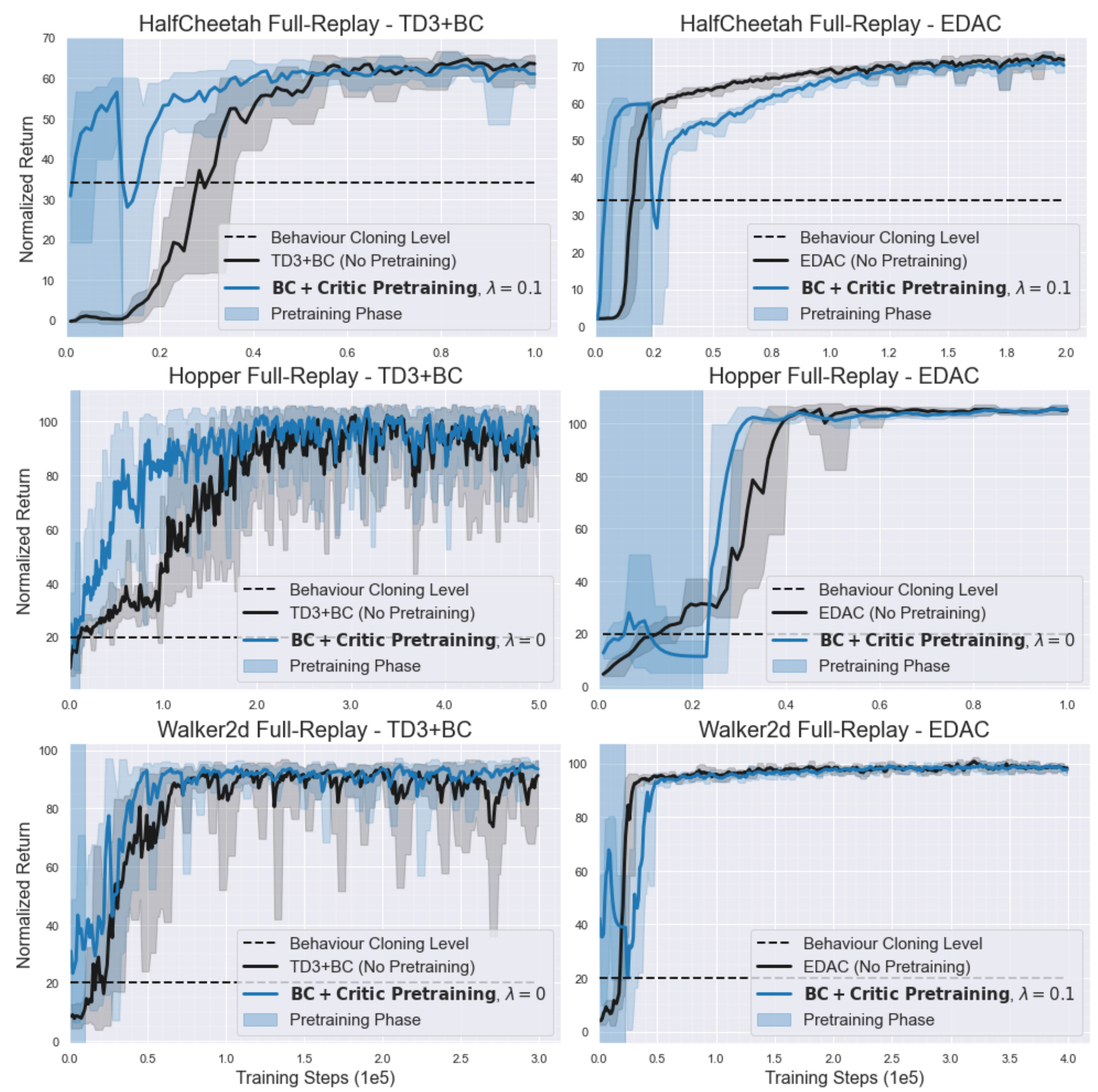}
  
  \caption[FR MuJoCo caption]%
  {Application of supervised pre-training to the \textit{full-replay} MuJoCo datasets. We generally see training efficiency gains similar to those observed in Figure \ref{fig:MuJoCo}. Plots show mean and standard deviation at each timestep for 3 independent runs.}
\vspace*{0mm}
\label{fig:FRMuJoCo}
\end{figure}

As with the \textit{medium-replay} datasets we see similar efficiency and stability gains to those observed for the \textit{medium} datasets in Figure \ref{fig:MuJoCo} (with the exception of the HalfCheetah environments). In some cases, the efficiency gains may be further improved by optimization of the pre-training duration and temporal-difference component $\lambda$ from those that were originally selected for the \textit{medium} datasets.

\newpage
\section{Investigation of Alternative Pre-training Method used by $Q$-Transformer}\label{app:qtrans}

As mentioned in Section \ref{sec:relatedwork}, $Q$-Transformer \citep{chebotar_q-transformer_2023} recently claimed improved training efficiency by incorporating MC values into the Bellman target via a max operation of the form $\max(MC, Q)$ (as originally proposed by \citet{wilcox_monte_2022}). We compare this objective to our combined actor and critic pre-training objective in Figure~\ref{fig:Qtrans} below.

\begin{figure}[h!]
\vspace*{-2mm}
\centering
  \includegraphics[width=\textwidth]{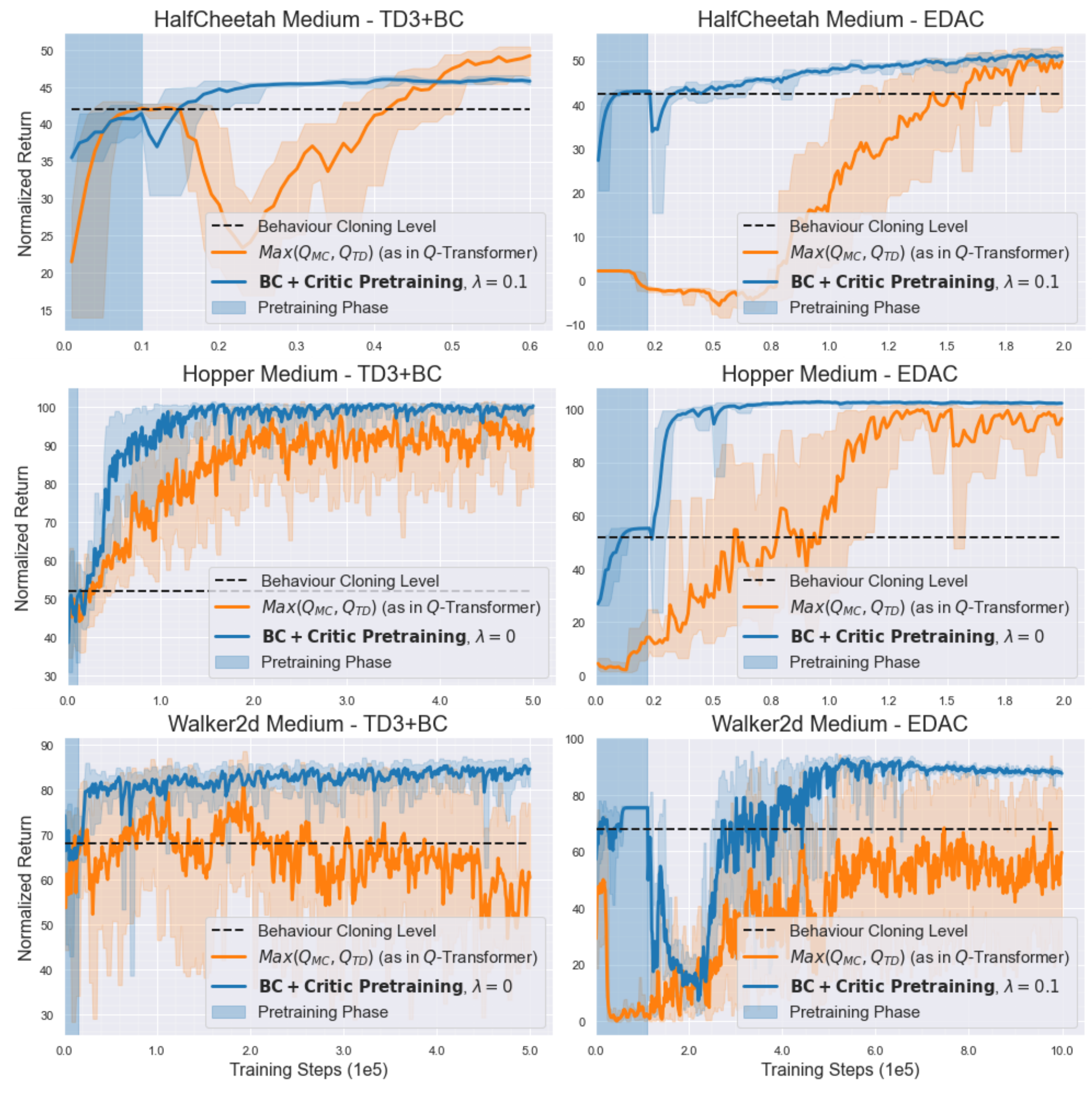}
  
  \caption[Qtrans caption]%
  {Comparison of the $Q$-Transformer training objective incorporating MC values via a $\max$ operation with our combined actor and critic pre-training. Plots show mean and standard deviation at each timestep for 3 independent runs.}
\vspace*{-3mm}
\label{fig:Qtrans}
\end{figure}

We find that our proposed combined actor and critic pre-training generally performs better than the $Q$-Transformer objective. Given the claimed performance benefit of this objective in $Q$-Transformer, we hypothesize that this relatively poor performance may be due to the variance in the MC values that is observed with these smaller scale datasets. In particular, in this smaller-scale offline setting, some MC values may be high variance which can prevent a clean transition from MC to TD ($Q$) values and prevent the temporal-difference loss from being properly optimized. We note that an effective hybrid approach might be to use this objective for pre-training, and end the pre-training when a certain threshold on the proportion of MC targets being selected by the max operation is reached, alleviating the ongoing impact of the MC values and the need for a pre-defined pre-training period. However, we leave a deeper investigation of this alternative objective to future work.

\newpage
\section{Extension to Data-Limited Adroit Environments}\label{sec:adroit}

In addition to our main MuJoCo experiments in Section \ref{sec:MuJoCo}, we consider the Adroit environments \citep{rajeswaran_learning_2018}, which are more complex and realistic tasks that require controlling a 24-DoF robotic hand to perform tasks such as aligning a pen, hammering a nail, opening a door, or relocating a ball. These form suitable environments for our approach, since the provided human demonstrations are much better than random behavior (and are therefore worth imitating initially) but can be improved upon by acting more efficiently, so it should be possible to improve on this imitation policy. However, the human datasets provided for these environments are very limited, consisting of just 25 trajectories of human demonstrations per task. The cloned datasets augment these trajectories with a behavior cloned policy to get a 50-50 mixture. 

\subsection{Motivation for Actor and Critic Regularization With Pre-Training}

In such data-limited settings however, off-policy algorithms often suffer from policy collapse as discussed in Section \ref{sec:hardprocedure}, since the actor or critic erroneously extrapolate out-of-distribution (OOD) of the offline data. Indeed, when we applied our pre-training approach from Section~\ref{sec:MuJoCo} to TD3+BC and EDAC on the Adroit environments, we found that the human-level pre-training performance often rapidly collapsed after pre-training. As we saw in Section \ref{sec:results} and Appendix \ref{app:Ablations}, consistency of both actor and critic are crucial for performance improvement. However, most offline RL methods apply regularization to only one of either the actor or the critic (as also noted by \citet{tarasov_revisiting_2023}). Motivated by this insight, we introduce two new algorithms that incorporate regularization on \textit{both} the actor and the critic to enable smooth performance improvement after pre-training in data-limited domains. First, we introduce TD3+BC+CQL, which combines the existing behavior cloning on the actor, with additional CQL regularization on the critic, to penalize large OOD Q-values. Second, we introduce EDAC+BC, which combines the existing uncertainty-based regularization on the critic, with additional behavior cloning on the actor, to penalize OOD actions.

\subsection{Implementation Details}\label{subsec:adroitimplementation}
For all baselines we use their benchmarked CORL implementations \citep{tarasov2022corl} and previously published hyperparameters where possible. Crucially, for the TD3+BC and EDAC baselines, we use the same regularization as TD3+BC+CQL and EDAC+BC for the existing regularization components. We train all algorithms for 300k updates (corresponding to up to 4 hours training time on our RTX2080 GPUs). For our novel algorithms, we pre-train for 200k steps, to provide ample time for supervised convergence. To fairly measure performance we then average online performance evaluated every 10000 offline updates between 200k and 300k updates, over 4 independent seeds. Full details of our procedure and hyperparameters are provided in Appendices \ref{app:adroitevaluation} and \ref{app:hyperparameters}.

\vspace{-0mm}
\begin{table}[h!]
\small
\centering
\addtolength{\tabcolsep}{-0.2em}
\begin{tabular}{ c | c c c c c c }
\hline
Env-Dataset & BC & CQL & TD3+BC & EDAC & \thead{\textbf{Pre-Trained}\\\textbf{TD3+BC+CQL}\\\textbf{(Ours)}} & \thead{\textbf{Pre-Trained}\\\textbf{EDAC+BC}\\\textbf{(Ours)}}\\
\hline
pen-human & $68.3 \pm 12.7$ & $51.1 \pm 3.8$ & $64.0 \pm 2.1$ & $9.2 \pm 1.6$ & $\mathbf{74.1 \pm 15.1}$ & $71.7 \pm 15.5$ \\
door-human & $0.6 \pm 0.7$ & $6.6 \pm 6.4$ & $0.1 \pm 0.1$ & $-0.2 \pm 0.1$ & $0.6 \pm 0.6$ & $\mathbf{13.2 \pm 3.8}$ \\
hammer-human & $9.6 \pm 4.4$ & $6.7 \pm 1.4$ & $5.6 \pm 2.5$ & $0.4 \pm 0.3$ & $10.6 \pm 6.1$ & $\mathbf{15.8 \pm 5.3}$ \\
relocate-human & $2.2 \pm 0.6$ & $0.6 \pm 0.2$ & $0.4 \pm 0.1$ & $0.5 \pm 0.1$ & $1.9 \pm 0.5$ & $\mathbf{3.6 \pm 0.5}$ \\
\hline
pen-cloned & $53.2 \pm 12.3$ & $50.3 \pm 3.9$ & $21.7 \pm 4.9$ & $13.6 \pm 12.5$ & $\mathbf{60.5 \pm 13.0}$ & $52.6 \pm 18.3$ \\
door-cloned & $0.3 \pm 0.5$ & $\mathbf{6.6 \pm 3.6}$ & $0.1 \pm 0.5$ & $-0.1 \pm 0.0$ & $0.0 \pm 0.3$ & $0.2 \pm 0.5$ \\
hammer-cloned & $3.6 \pm 1.4$ & $4.0 \pm 1.5$ & $2.8 \pm 2.0$ & $0.7 \pm 0.1$ & $1.4 \pm 0.6$ & $\mathbf{8.3 \pm 6.4}$ \\
relocate-cloned & $0.1 \pm 0.0$ & $0.2 \pm 0.1$ & $-0.1 \pm 0.1$ & $0.2 \pm 0.1$ & $0.1 \pm 0.0$ & $0.1 \pm 0.1$ \\
\hline
\end{tabular}
\vspace{1mm}
\caption{Normalized average returns ($\pm$ standard deviation), by averaging performance between 200k and 300k updates over 4 independent seeds. Our combined algorithms ensure that \textit{both} the actor and critic are regularized to stay close to the behavior policy after pre-training, often giving greater performance than the component algorithms when learning from limited demonstrations.}
\label{tab:Adroit}
\vspace{-2mm}
\end{table}

\subsection{Results and Analysis}
We find that behavior cloning (BC) provides a strong baseline, notably greatly outperforming previously quoted BC performances on these environments due to our addition of LayerNorm. We see that CQL and EDAC (with LayerNorm) perform reasonably with this evaluation approach using the same hyperparameters as specified in the original papers \citep{kumar_conservative_2020, an_uncertainty-based_2021}, although may benefit from a greater training budget. The original TD3+BC paper did not consider the Adroit environments, but we see comparable performances with relatively strong behavior cloning regularization (tuned with $\alpha=1$ for the pen environments and $\alpha=0.1$ for other environments). However, our additions of CQL regularization to TD3+BC and BC regularization to EDAC both generally lead to improved performance in these environments. In fact, we find that our new hybrid algorithms perform similarly using this evaluation procedure even without pre-training, due to their strong inherent regularization, as shown and discussed further in Appendix \ref{app:adroitablation}. Performance plots in these environments for the human datasets are provided in Appendix~\ref{app:adroitplots}.

\subsection{Key Hyperparameters for Adroit Experiments}\label{app:hyperparameters}
The Adroit experiments were performed and evaluated as described in Section 5 of the main paper, with more detail provided below in Appendix \ref{app:adroitevaluation}. Key hyperparameters for each algorithm are provided in Table~3 below. Full hyperparameter configurations (including those not provided below) are available in the config files of the provided CORL codebase \citep{tarasov2022corl}. Where possible, hyperparameters were chosen to match previously published values for the Adroit environments, and otherwise their default implementation values. Crucially, where hyperparameters are shared between algorithms (such as between TD3+BC and TD3+BC+CQL) they were chosen to be equal, to investigate whether additional regularization on the critic/actor can improve over the exisiting tuned regularization on the actor/critic alone.

To incorporate our additional regularization losses which may be of different scales to the existing losses, we utilize the normalization strategy described in TD3+BC \citep{fujimoto_minimalist_2021}. Namely for primary loss $\alpha$ and additional auxiliary loss $\beta$ which may be of a different scale, we combine them as follows to more evenly balance the losses throughout training:
\begin{equation}
    \mathcal{L} = \alpha/|\alpha| + c \ \beta/|\beta|
\end{equation}
where $|\cdot|$ denotes the magnitude of the gradient-detached loss and $c$ is the regularization coefficient referred to as CQL/BC-regularizer provided in Table~3.

Finally, we note that for the behavior cloning baseline and for the behavior cloning regularization in both TD3+BC(+CQL) and EDAC+BC we use `hard' behavior cloning (using a mean-squared error objective). In the case of the BC baseline and EDAC+BC it would be possible to use `soft' behavior cloning as in Equation~6, but we found in both cases `hard' behavior cloning (using the sampled action from the Gaussian policy for EDAC) performed much better. However we still use `soft' behavior cloning for pre-training EDAC+BC to maintain the policy entropy in pre-training.
\begin{table}[h!]
\centering
\caption{Adroit Experiments Key Hyperparameters}\label{tab:hyperparameters}
\begin{center}
\scalebox{0.93}{
\begin{tabular}{cccc}
\bf ALGORITHM&\bf TASK&\bf PARAMETER&\bf VALUE
\\ \hline \\
BC & All & BC Objective & MSE \\
CQL & All & n-actions & 10 \\
CQL & All & Temperature & 1.0 \\
TD3+BC(+CQL) & Pen & $\alpha$ & 1.0 \\
TD3+BC(+CQL) & Door/Hammer/Relocate & $\alpha$ & 0.1 \\
TD3+BC+CQL & Pen & CQL-regularizer & 1.0 \\
TD3+BC+CQL & Door/Hammer/Relocate & CQL-regularizer & 10.0 \\
TD3+BC+CQL & All & n-actions & 10 \\
TD3+BC+CQL & All & Temperature & 1.0 \\
TD3+BC+CQL & All & Pre-training $\lambda$ & 0 \\
EDAC(+BC) & Pen & N (num critics) & 20 \\
EDAC(+BC) & Pen (Human) & $\eta$ & 1000 \\
EDAC(+BC) & Pen (Cloned) & $\eta$ & 10 \\
EDAC(+BC) & Door/Hammer/Relocate & N (num critics) & 50 \\
EDAC(+BC) & Door/Hammer/Relocate & $\eta$ & 200 \\
EDAC+BC & All & BC Objective & MSE \\
EDAC+BC & All & BC-regularizer & 1.0 \\
EDAC+BC & All & Pre-training $\lambda$ & 0 \\
\end{tabular}}
\end{center}
\end{table}

\newpage
\subsection{Additional Discussion of Evaluation Procedure for Adroit Experiements}
\label{app:adroitevaluation}

As described in Section \ref{subsec:adroitimplementation}, we train all algorithms using the standard hyperparameters provided for 300k steps, and evaluate their performance by evaluating the agent performance every 10k steps between 200k and 300k steps. For the algorithms including pre-training, we pre-train for 200k steps to allow more than sufficient convergence of the pre-training stage (using $\lambda=0$). There are two motivations for this evaluation procedure. Firstly, we would like to measure the performance of the algorithms after a relatively short training time (i.e. reduced number of training steps) to provide a quantitative measure of the performance efficiency. Secondly, since the performance of the offline RL algorithms considered on this benchmark (and in general) are very unstable, online performance varies significantly between random and human performance during training. Therefore, to reduce the variance of the results and to incorporate performance stability, we do not take a single (or best) checkpoint, but rather average the performance of checkpoints taken every 10k steps between 200k and 300k training steps, each evaluated online for 10 episodes. This is more representative of real world performance off offline RL algorithms where the online performance may not be possible to evaluate during training.

One additional consideration for evaluation resulted from the fact that we noticed that, aside from the \textit{pen} environment, the human demonstrations in the human datasets were much longer than the truncation limit of the online truncation limit of the environments. The truncation limit for all environments (for the v1 environments other than \textit{pen}) is set to 200 timesteps, while the maximum demonstration lengths are 300, 624 and 527 timesteps for the \textit{Door}, \textit{Hammer} and \textit{Relocate} environments respectively. This partially explains the poor performance of previous algorithms on the non-\textit{pen} environments in this benchmark \citep{an_uncertainty-based_2021}, since the environment does not allow sufficient time to receive reward for successfully imitating the demonstrated behavior, such as opening the door or hammering the nail, before truncating the episode. Therefore we adjust the truncation limit in these environments to match the maximum demonstration lengths. To compute the standard human normalized score (defined as $(\mathrm{agent\_score} - \mathrm{random\_score})/(\mathrm{human\_score} - \mathrm{random\_score})$, we maintain the same maximal human scores as provided in the D4RL benchmark, but adjust the minimal random scores by running the independent uniform random policy in the environments for the new truncation time limits. This gives rise to the new environment evaluation variables provided below in Table \ref{tab:eval}. Crucially, despite this improved evaluation procedure, we evaluate all algorithms considered using this procedure in an identical manner, to provide a fair comparison of algorithm performance that takes into account both efficiency and stability.

\begin{table}[h!]
\centering
\caption{Adroit Environment Evaluation Parameters}\label{tab:eval}
\begin{center}
\begin{tabular}{cccc}
\bf ENV&\bf \thead{NEW\\TIMESTEP\\LIMIT}&\bf \thead{NEW MIN \\(RANDOM)\\SCORES}&\bf \thead{MAX\\(HUMAN)\\SCORES}
\\ \hline 
{pen}&$100$&$-162.09$&$3076.83$\\
{door}&$300$&$-84.52$&$2880.57$\\
{hammer}&$624$&$-856.83$&$12794.13$\\
{relocate}&$527$&$-37.95$&$4233.88$\\
\end{tabular}
\end{center}
\end{table}

\subsection{Pre-Training Ablation on Adroit Environments}\label{app:adroitablation}
 We ablate the pre-training stage of our new hybrid algorithms, TD3+BC+CQL and EDAC+BC, on the Adroit environment below in Table \ref{tab:ptablation}.

\begin{table}[h!]
\centering
\caption{Pre-training Ablation for Our Hybrid Algorithms on Adroit Environments.}\label{tab:ptablation}
\begin{center}
\begin{tabular}{l|ll|ll}
\hline
 & \thead{TD3-BC-CQL} & \thead{Pre-trained\\TD3-BC-CQL} & \thead{EDAC-BC} & \thead{Pre-trained\\EDAC-BC} \\
\hline
pen-human & $79.7 \pm 10.1$ & $74.1 \pm 15.1$ & $\mathbf{83.8 \pm 14.8}$ & $71.7 \pm 15.5$ \\
door-human & $0.6 \pm 0.3$ & $0.6 \pm 0.6$ & $0.9 \pm 1.0$ & $\mathbf{13.2 \pm 3.8}$ \\
hammer-human & $13.2 \pm 9.7$ & $10.6 \pm 6.1$ & $9.3 \pm 3.8$ & $\mathbf{15.8 \pm 5.3}$ \\
relocate-human & $2.9 \pm 1.4$ & $1.9 \pm 0.5$ & $2.1 \pm 0.7$ & $\mathbf{3.6 \pm 0.5}$ \\
\hline
pen-cloned & $\mathbf{63.7 \pm 12.5}$ & $60.5 \pm 13.0$ & $56.3 \pm 21.3$ & $52.6 \pm 18.3$ \\
door-cloned & $0.1 \pm 0.4$ & $0.0 \pm 0.3$ & $\mathbf{0.8 \pm 0.7}$ & $0.2 \pm 0.5$ \\
hammer-cloned & $2.2 \pm 0.7$ & $1.4 \pm 0.6$ & $4.9 \pm 2.6$ & $\mathbf{8.3 \pm 6.4}$ \\
relocate-cloned & $0.0 \pm 0.1$ & $\mathbf{0.1 \pm 0.0}$ & $0.0 \pm 0.0$ & $0.1 \pm 0.1$ \\
\hline
\end{tabular}
\end{center}
\end{table}

We see that both versions of each algorithm (with and without pre-training) perform similarly, although there appears to be a slight benefit from pre-training for EDAC+BC in some environments. This is likely partially just because there is little efficiency gain to be made from pre-training in this data-limited setting, and partly because the combined regularization is sufficient to keep the actor and critic consistent with each other and with the data distribution. Also both of the additional regularization components (behavior cloning and CQL-style regularization) do not rely on temporal difference bootstrapping and therefore have similar efficiency to supervised learning and reduce the benefit of pre-training. However, a shorter pre-training period may demonstrate greater benefits from pre-training (since 200k steps is really more than required, providing additional time for non-pre-trained versions to learn), along with increased dataset size. Importantly, these hybrid algorithms, motivated by our pre-training approach, still demonstrate promising improvements in performance relative to the component algorithms in this data-limited setting, as demonstrated in Table \ref{tab:Adroit}. 

We also note that even with our substantial efforts to make our evaluation procedure as `fair' as possible between algorithms (as described in Appendices \ref{app:hyperparameters} and \ref{app:adroitevaluation}) the variances (and therefore confidence intervals) of these Adroit results are still non-negligible due to the nature of the limited human data, the stochastic starting states of the environment, the high variance algorithms used, and our limited computation resources. However, our aim is not to show that any one algorithm is `best', as this is dependent on a wide range of factors and is often entirely infeasible in general \citep{patterson_empirical_2023}. Indeed, the performance of these algorithms is generally comparable (as intuitively might be expected given the same limited behavior data), and significantly improving on the behavior policy is challenging. Rather, our results on this benchmark aim to demonstrate the idea that if the performance of an algorithm collapses after pre-training (or more generally, imitation learning gives rise to non-negligible performance but off-policy RL does not), this can be mitigated by introducing additional regularization towards the behavior policy, and it is often more effective to regularize \textit{both} the actor and the critic rather than just one of these components, as in Table \ref{tab:Adroit}. 

\newpage
\subsection{Performance Plots for Adroit Experiments}\label{app:adroitplots}

\begin{figure}[h!]
\vspace*{0mm}
\centering
  \includegraphics[width=1.0\textwidth]{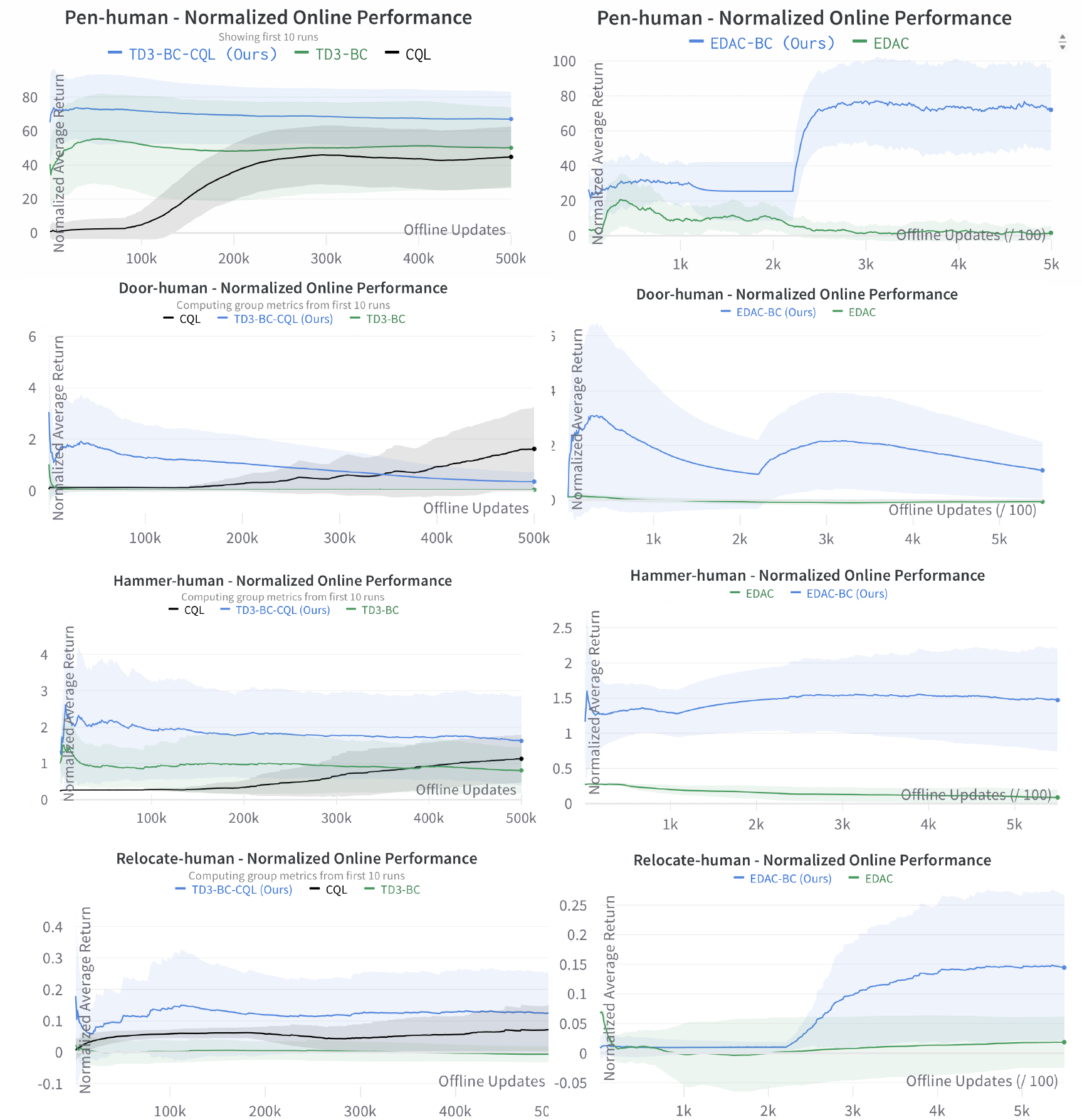}
  
  \caption[Adroit plots caption]%
  {Training plots for Adroit environments using the human datasets. As described in Section 5, we introduce additional combined regularization, giving rise to novel hybrid algorithms TD3+BC+CQL and EDAC+BC, to prevent performance collapse after pre-training. We find that combining actor and critic regularization leads to better performance than equivalent actor or critic regularization alone (regularization hyperparameters provided in Table 3). However, we see that in many of these data-limited environments, subsequent off-policy reinforcement learning is not able to improve upon the initial pre-training performance corresponding to imitation learning (with LayerNorm). Plots show mean and standard deviation at each timestep for 4 independent runs.}
\vspace*{0mm}
\label{fig:AdroitPlots}
\end{figure}
\newpage

\end{document}